\PassOptionsToPackage{table}{xcolor}
\documentclass[sigconf]{acmart}
\AtBeginDocument{%
  }

\usepackage{multirow}
\usepackage{subfig}
\usepackage{xcolor}
\usepackage{enumitem}
\usepackage{pifont}

\copyrightyear{2026}
\acmYear{2026}
\setcopyright{cc}
\setcctype{by}
\acmConference[MM '26]{Proceedings of the 34th ACM International Conference on Multimedia}{November 10--14, 2026}{Rio de Janeiro, Brazil}
\acmBooktitle{Proceedings of the 34th ACM International Conference on Multimedia (MM '26), November 10--14, 2026, Rio de Janeiro, Brazil}
\acmDOI{10.1145/3767308.3838634}
\acmISBN{979-8-4007-2213-4/2026/11}

\begin{document}

\title{AGIDefect-4K: A Richly Annotated Dataset for AI-Generated Image Defect Detection, Localization and Explanation}

\author{Xiangfei Sheng}
\orcid{0009-0004-8468-1970}
\authornotemark[1]
\affiliation{%
  \institution{Xidian University}
  \city{Xi'an}
  \state{Shaanxi}
  \country{China}
}
\email{xiangfeisheng@gmail.com}

\author{Weidong Zou}
\orcid{0009-0001-4284-6042}
\authornote{Both authors contributed equally to this research.}
\affiliation{%
  \institution{Xidian University}
  \city{Xi'an}
  \state{Shaanxi}
  \country{China}
}
\email{zouweidong@stu.xidian.edu.cn}

\author{Tianjiao Gu}
\orcid{0009-0005-5690-7909}
\affiliation{%
  \institution{Xidian University}
  \city{Xi'an}
  \state{Shaanxi}
  \country{China}
}
\email{gutianjiao@stu.xidian.edu.cn}

\author{Zhichao Yang}
\orcid{0009-0008-3398-1286}
\affiliation{%
  \institution{Xidian University}
  \city{Xi'an}
  \state{Shaanxi}
  \country{China}
}
\email{yangzhichao@stu.xidian.edu.cn}

\author{Pengfei Chen}
\orcid{0000-0002-0509-3782}
\affiliation{%
  \institution{Xidian University}
  \city{Xi'an}
  \state{Shaanxi}
  \country{China}
}
\email{chenpengfei@xidian.edu.cn}

\author{Leida Li}
\orcid{0000-0001-9069-8796}
\authornote{Corresponding author.}
\affiliation{%
  \institution{Xidian University}
  \city{Xi'an}
  \state{Shaanxi}
  \country{China}
}
\email{ldli@xidian.edu.cn}

\renewcommand{\shortauthors}{Xiangfei Sheng et al.}

\begin{abstract}

Generative AI can now produce highly realistic images, yet current models still exhibit subtle but critical defects that undermine their reliability. While existing AI-generated image (AGI) evaluation benchmarks have made notable progress, comprehensive AGI defect diagnosis remains underexplored. To bridge this gap, we introduce \textbf{AGIDefect-4K}, a richly annotated dataset of \textbf{4,000} images from \textbf{15} state-of-the-art generative models spanning both open-source and closed-source systems. AGIDefect-4K features \textbf{hierarchical defect annotations}: (1) detection labels identifying whether defects exist, (2) pixel-level segmentation masks localizing defective regions, and (3) detailed textual explanations characterizing defect types and their perceptual impact. Each image is further annotated with an overall quality score. Building on this, we present \textbf{AGIDA} (\textbf{AGI} \textbf{D}efect \textbf{A}ssistant), a baseline framework leveraging Multimodal Large Language Models (MLLMs) for joint defect detection, localization, explanation, and quality prediction. Comprehensive benchmarking on AGIDefect-4K reveals that AGI defect understanding remains challenging, underscoring the value of this dataset. The dataset is publicly available at \url{https://github.com/sxfly99/AGIDefect-4K}.

\end{abstract}

\ccsdesc[500]{Human-centered computing~Visualization design and evaluation methods}

\keywords{AI-Generated Images, Defect Detection, Quality Assessment}

\begin{teaserfigure}
\centering
  \includegraphics[width=0.93\textwidth]{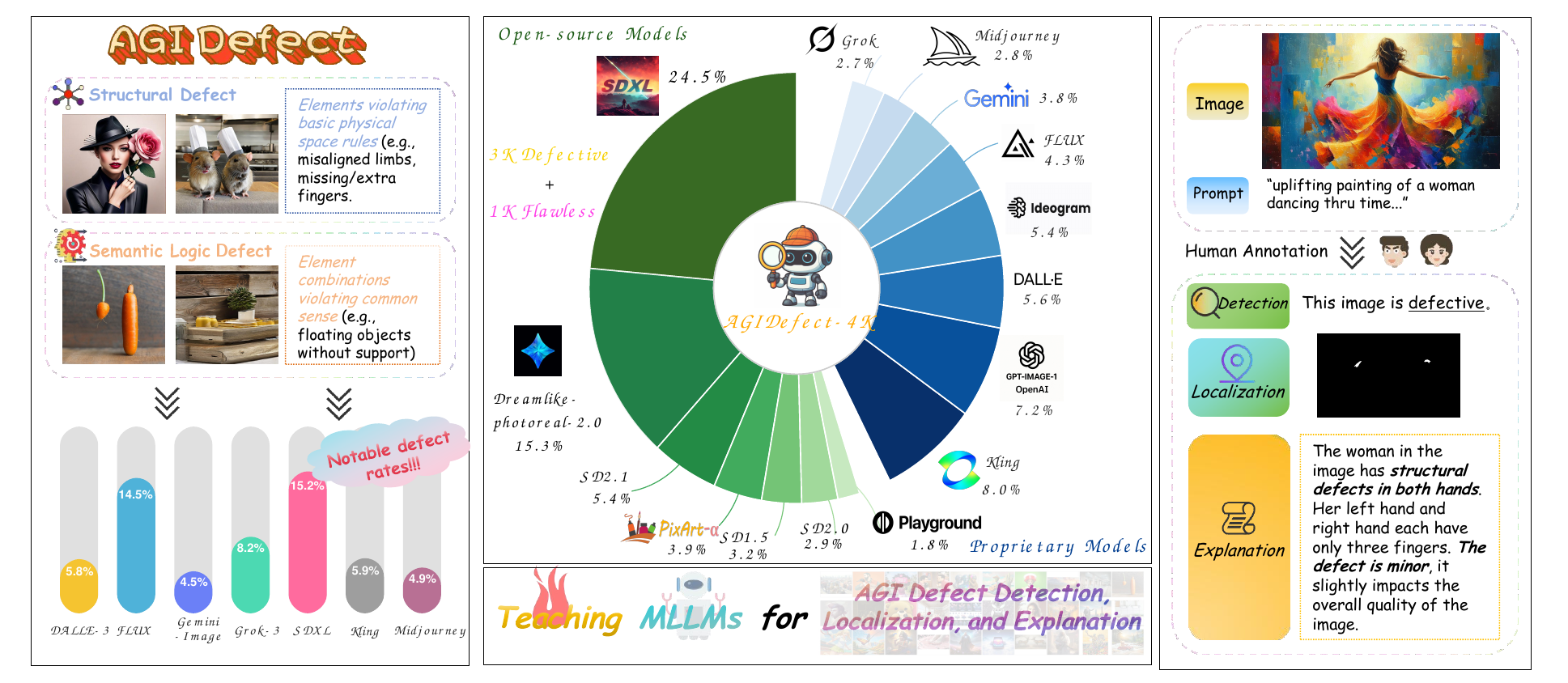}
  \Description{Overview figure showing motivation, dataset, and annotation framework.}
\caption{\textbf{Overview of our approach.} \textit{Left:} Illustration of our motivation. Even state-of-the-art generative models exhibit notable defect rates. \textit{Middle:} AGIDefect-4K dataset comprising images from 15 diverse generative models. \textit{Right:} Our hierarchical annotation framework: detection (binary classification), localization (segmentation masks), and explanation (detailed textual descriptions), enabling comprehensive defect detection through multimodal large language models.}
\label{fig:teaser}
\end{teaserfigure}

\maketitle

\section{Introduction}

The unprecedented success of generative artificial intelligence has revolutionized digital content creation, enabling the synthesis of photorealistic images with remarkable fidelity~\cite{everypixel2024,podell2023sdxl}. Despite these achievements, contemporary AI-generated images (AGIs) remain susceptible to critical defects that compromise their reliability and authenticity. These defects manifest predominantly as \textit{structural defects} (e.g., anatomical errors, misaligned limbs, impossible geometries) and \textit{semantic logic defects} (e.g., physically implausible object relationships, floating objects without support). As shown in Figure~\ref{fig:teaser}, our annotation statistics reveal that even state-of-the-art generative models exhibit defect rates ranging from 4.5\% to 15.2\%, underscoring the urgent need for systematic AGI defect detection—not only to enable defect identification and quality evaluation, but also to provide essential feedback for iterative model improvement and trustworthy real-world deployment.

Recent advances in AGI evaluation have recognized generation defects as a significant quality factor~\cite{abench,agin,Evalmuse,Q-eval-100k}. However, most methodologies treat defects as an ancillary factor within holistic evaluation frameworks rather than focal points for dedicated analysis. While recent efforts have begun to explore defect-aware evaluation through heatmap-based approaches~\cite{RichHF-18K, yang2025heie}, comprehensive and systematic defect diagnosis remains underexplored. This raises a pivotal question: 

\textit{How should we design annotations to enable comprehensive AGI defect diagnosis?}

We posit that effective defect analysis necessitates a hierarchical approach spanning three critical dimensions: detection (identifying defect presence), localization (pinpointing affected regions), and explanation (characterizing defect nature and impact). This multi-tiered structure mirrors human cognitive processes in visual perception—we instinctively notice anomalies, locate them spatially, and then comprehend their implications. Complemented by an overall quality score that captures the perceptual impact of defects, these annotations form the foundation of our work. Our contributions are as follows:

\begin{itemize}[leftmargin=*]
\item \textbf{Dataset.} We introduce \textbf{AGIDefect-4K}, comprising 4,000 images (3,000 defective and 1,000 flawless) generated by 15 state-of-the-art models spanning both open-source and closed-source systems. Each image is rigorously annotated by at least three human experts with hierarchical defect annotations (detection, localization, and explanation) and overall quality scores, with double-checking procedures ensuring high annotation reliability.

\item \textbf{Baseline.} We present \textbf{AGIDA} (AGI Defect Assistant), a baseline framework leveraging MLLMs for joint defect detection, localization, explanation, and quality prediction.

\item \textbf{Analysis.} We benchmark a wide range of models on AGIDefect-4K, including MLLMs, defect detection models, and quality assessment models, revealing challenges in defect understanding and highlighting the potential of defect analysis for advancing AGI quality assessment.

\end{itemize}

\begin{figure*}[t]
  \centering
  \includegraphics[width=0.7\textwidth ]{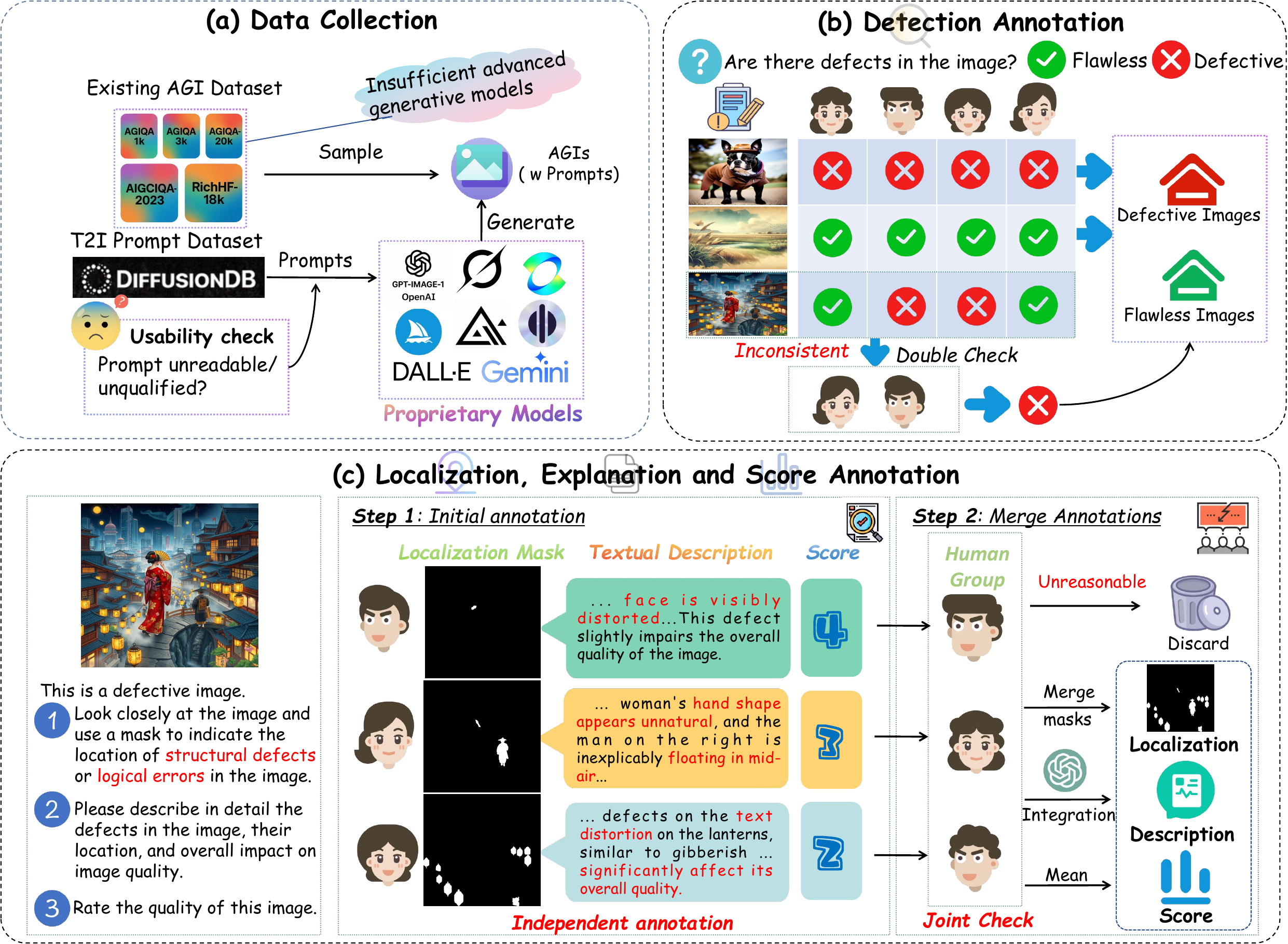}
\caption{Overview of AGIDefect-4K dataset construction pipeline.}
\Description{The three-stage construction pipeline of AGIDefect-4K. First, image-prompt pairs are sampled from existing AI-generated image datasets, while additional images are generated from filtered DiffusionDB prompts using proprietary models. Second, multiple annotators independently classify each image as defective or flawless, and disputed cases undergo joint review. Third, defective images receive independently annotated segmentation masks, textual explanations, and quality scores. An expert panel then discards invalid annotations, merges valid masks and descriptions, and averages the quality scores.}
\label{fig:construction}

\end{figure*}

\begin{table}[t]
\centering
\caption{Comparison with existing AGI evaluation datasets. \textbf{Det.}: defect detection. \textbf{Loc.}: defect localization (H: heatmap, M: mask). \textbf{Exp.}: textual explanation. \textbf{QS}: quality score.}
\label{tab:dataset_comparison}
\footnotesize
\begin{tabular}{@{}lrccccc@{}}
\toprule
\textbf{Dataset} & \textbf{\#Img} & \textbf{Pub} & \textbf{Det.} & \textbf{Loc.} & \textbf{Exp.} & \textbf{QS} \\
\midrule
AGIQA-3K~\cite{AGIQA-3k} & 2,982 & TCSVT'24 & \ding{55} & \ding{55} & \ding{55} & \ding{51} \\
RichHF-18K~\cite{RichHF-18K} & 18,000 & CVPR'24 & \ding{51} & \textbf{H} & \ding{55} & \ding{51} \\
A-Bench~\cite{abench} & 2,864 & ICLR'25 & \ding{55} & \ding{55} & \ding{51} & \ding{55} \\
Q-Eval-100K~\cite{Q-eval-100k} & 100,000 & CVPR'25 & \ding{55} & \ding{55} & \ding{55} & \ding{51} \\
AIGI-VC~\cite{tian2025aigivc} & 2,500 & AAAI'25 & \ding{55} & \ding{55} & \ding{51} & \ding{51} \\
EvalMuse-40K~\cite{Evalmuse} & 40,000 & AAAI'26 & \ding{51} & \ding{55} & \ding{55} & \ding{55} \\

\midrule
\rowcolor{blue!5}
\textbf{AGIDefect-4K (Ours)} & 4,000 & MM'26 & \ding{51} & \textbf{M} & \ding{51} & \ding{51} \\
\bottomrule
\end{tabular}
\vspace{-3ex}
\end{table}

\section{Related Work}

As summarized in Table~\ref{tab:dataset_comparison}, the evaluation of AI-generated images (AGIs) has evolved from holistic scoring to more granular feedback. Early benchmarks such as AGIQA-1K~\cite{AGIQA-1K}, AGIQA-3K~\cite{AGIQA-3k}, and AIGCIQA2023~\cite{aigciqa2023} established perceptual quality assessment through numerical ratings, with AIGCIQA2023 incorporating authenticity metrics. Subsequent large-scale efforts including AGIQA-20K~\cite{AGIQA-20K} and Q-Eval-100K~\cite{Q-eval-100k} expanded the evaluation scale. AIGI-VC~\cite{tian2025aigivc} extended evaluation to visual communication scenarios with preference descriptions, while LongT2IBench~\cite{yang2026longt2ibench} introduced graph-structured annotations for long text-to-image generation. Related studies have explored fine-grained image quality and aesthetics assessment~\cite{sheng2023aesclip,sheng2025aesprompt,sheng2026fine,sheng2026tuningiqa,yang2024semantics,yang2025language,yang2026fine}. While these works acknowledged AI artifacts as quality factors, score-based approaches inherently lack the granularity to specify defect locations or characteristics.

Related deepfake detection studies have investigated generalization-preserved continual learning, uncertainty-guided expert selection, and vision-language semantics~\cite{zhang2025generalization,zhang2025choose,zhu2026unleashing}. RichHF-18K~\cite{RichHF-18K} introduced artifact detection labels and heatmap-based localization, and HEIE~\cite{yang2025heie} further leveraged such annotations to build an MLLM-based evaluator. EvalMuse-40K~\cite{Evalmuse} categorized structural problems through predefined taxonomies, while A-Bench~\cite{abench} employed question-answer pairs to evaluate quality perception and semantic understanding. Despite these advances, a dedicated benchmark with comprehensive defect annotations remains underexplored. AGIDefect-4K distinguishes itself by providing hierarchical annotations that unify defect detection, pixel-level localization, and detailed textual explanation, complemented by overall quality scores.

\section{AGIDefect-4K Dataset}

\subsection{Basic Principles}

The construction of AGIDefect-4K, illustrated in Figure~\ref{fig:construction}, follows two fundamental principles.

\textbf{Principle 1: Maximizing diversity} by collecting images from existing AGI datasets and generating additional content using multiple state-of-the-art proprietary models, ensuring comprehensive representation across generative architectures and failure modes.

\textbf{Principle 2: Ensuring annotation reliability} through a rigorous two-stage protocol consisting of \textit{independent annotation} followed by \textit{joint verification}, carried out by a team of 20 trained expert annotators. Detailed definitions of structural and semantic logic defects are provided in the \textit{supplementary material}.

\begin{figure}[!tbp]
    \centering
    \subfloat[\label{fig:score}]{%
        \includegraphics[width=0.48\columnwidth]{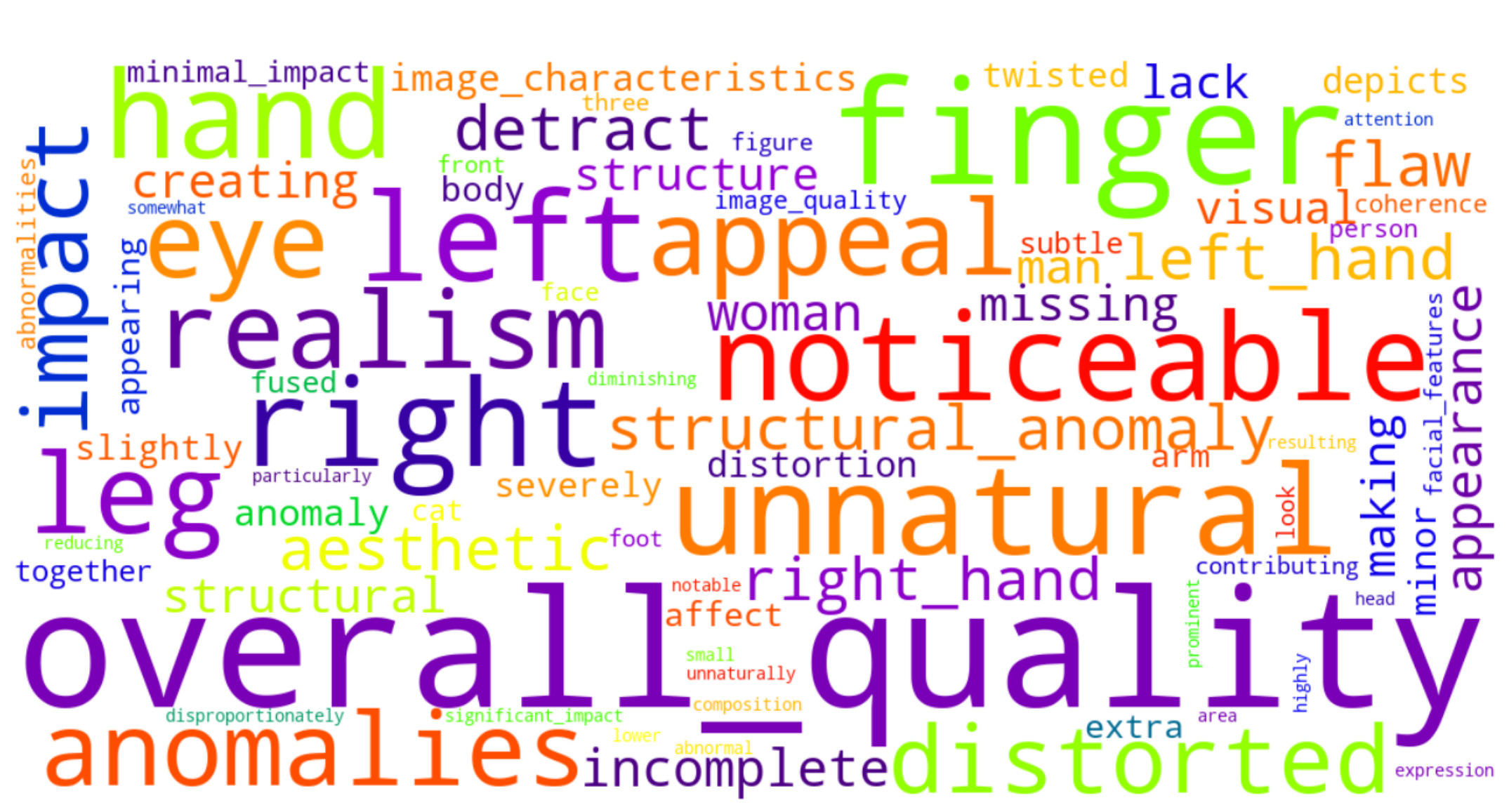}%
    }
    \hfill
    \subfloat[\label{fig:diff}]{%
        \includegraphics[width=0.48\columnwidth]{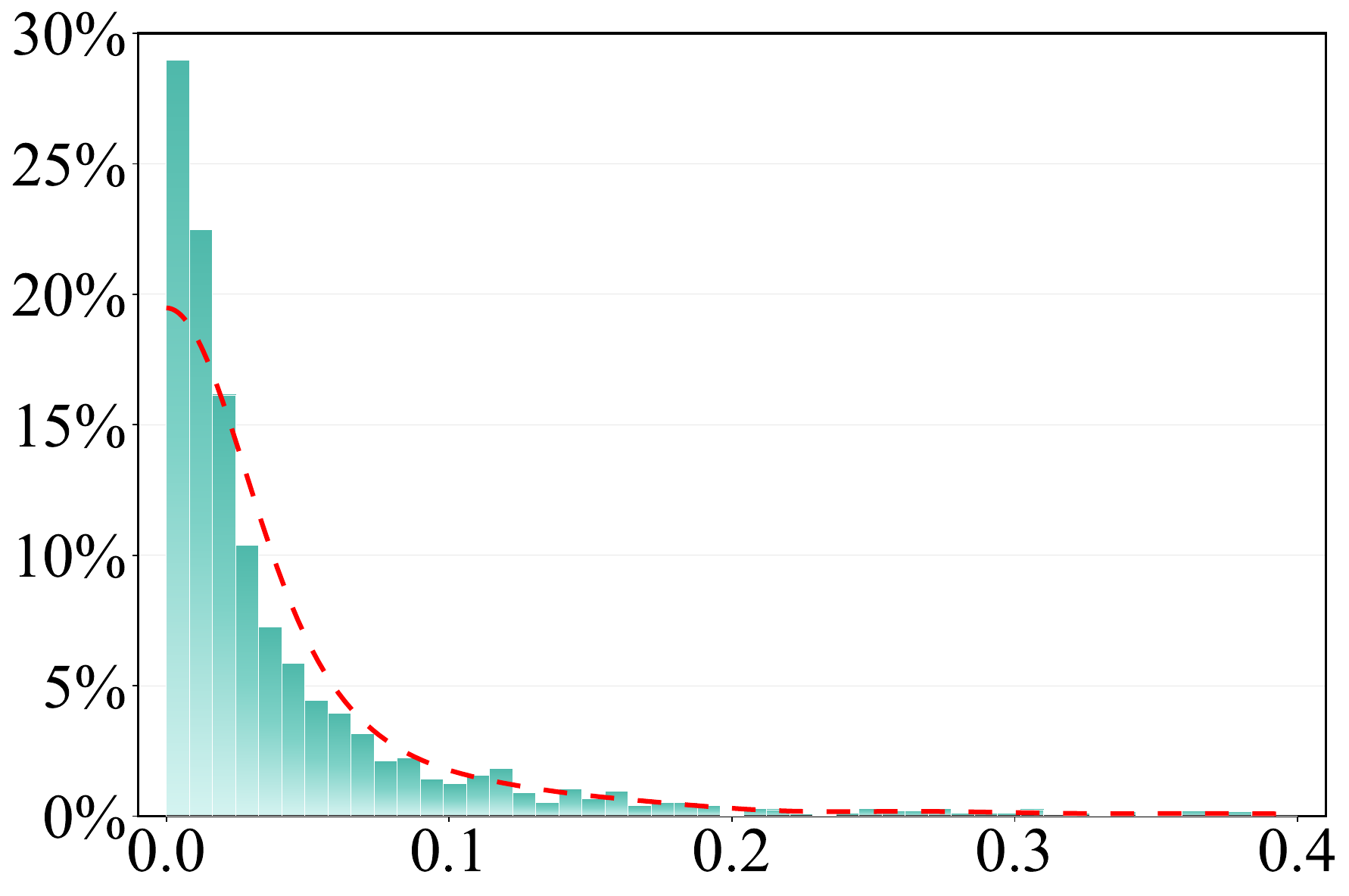}%
    }\\
    \subfloat[\label{fig:ele}]{%
        \includegraphics[width=0.48\columnwidth]{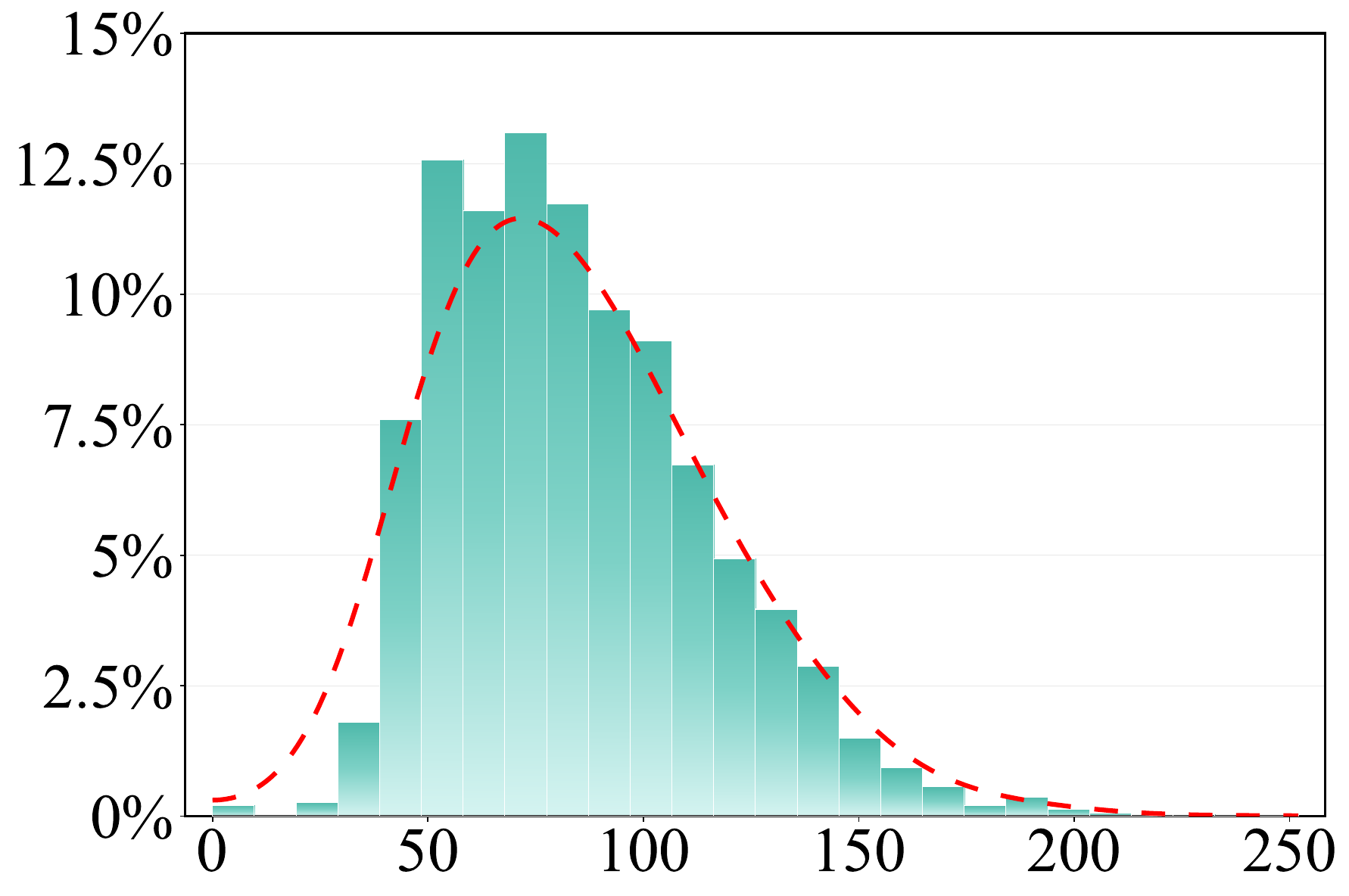}%
    }
    \hfill
    \subfloat[\label{fig:new}]{%
        \includegraphics[width=0.48\columnwidth]{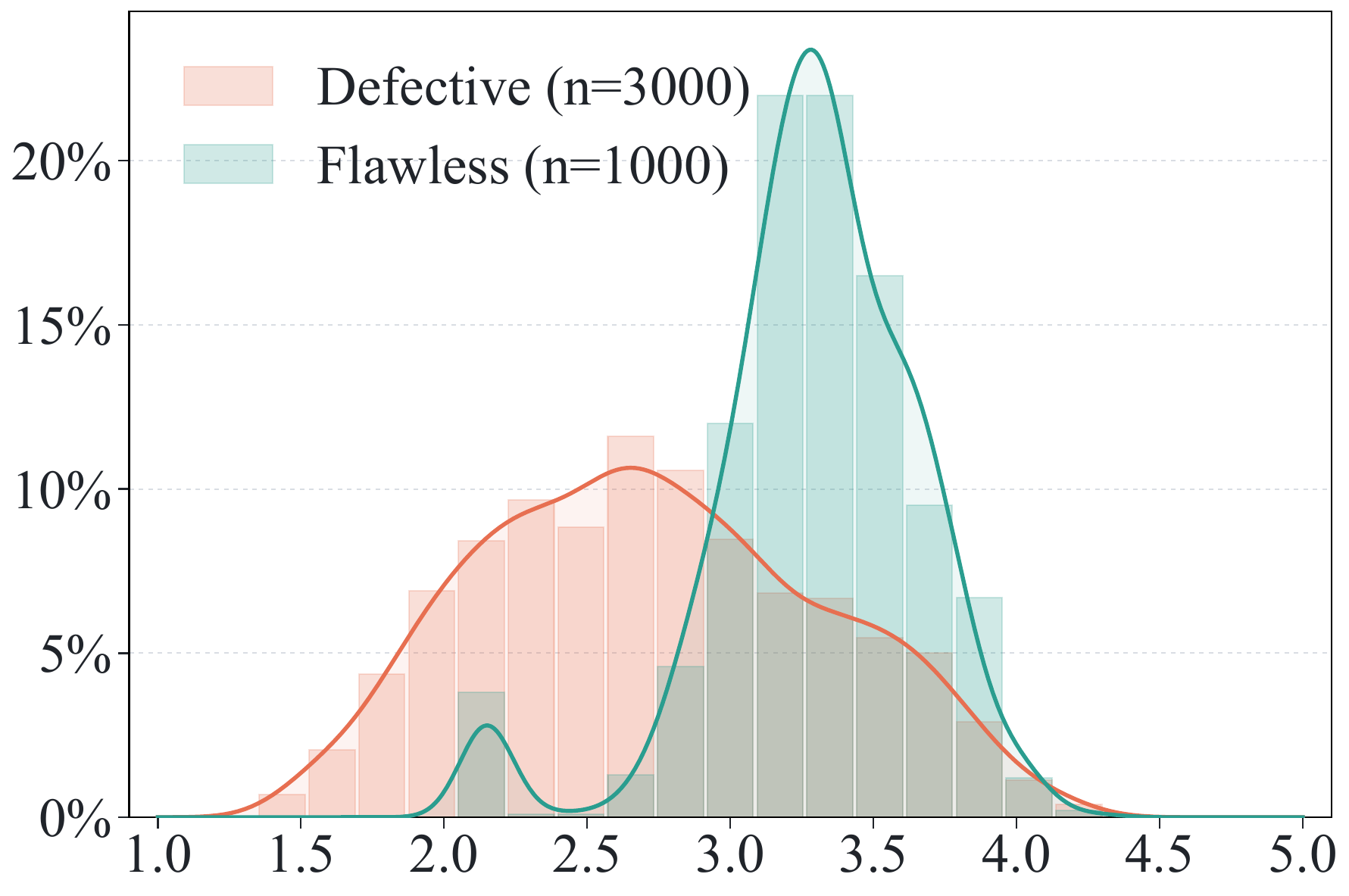}%
    }
    
    \caption{\textbf{Statistical analysis of AGIDefect-4K.} (a) Word cloud of textual descriptions. (b) Distribution of defect mask sizes as a percentage of total image area. (c) Token length distribution of defect descriptions. (d) Quality score distributions for defective and flawless images.}
\Description{Four statistical summaries of AGIDefect-4K. The word cloud highlights frequent concepts in the defect explanations, including overall quality, noticeable, unnatural, finger, hand, eye, and structural anomaly. The defect-mask area distribution is strongly right-skewed, showing that most defects occupy a small proportion of an image. Description lengths concentrate around 50 to 100 tokens with a long tail approaching 200 tokens. Quality scores for defective images are lower and more broadly distributed than those for flawless images, whose scores concentrate around 3.3.}
    \label{fig:statistics}
\vspace{-3ex}
\end{figure}

\begin{figure}[t]
\centering
\includegraphics[width=0.95\columnwidth]{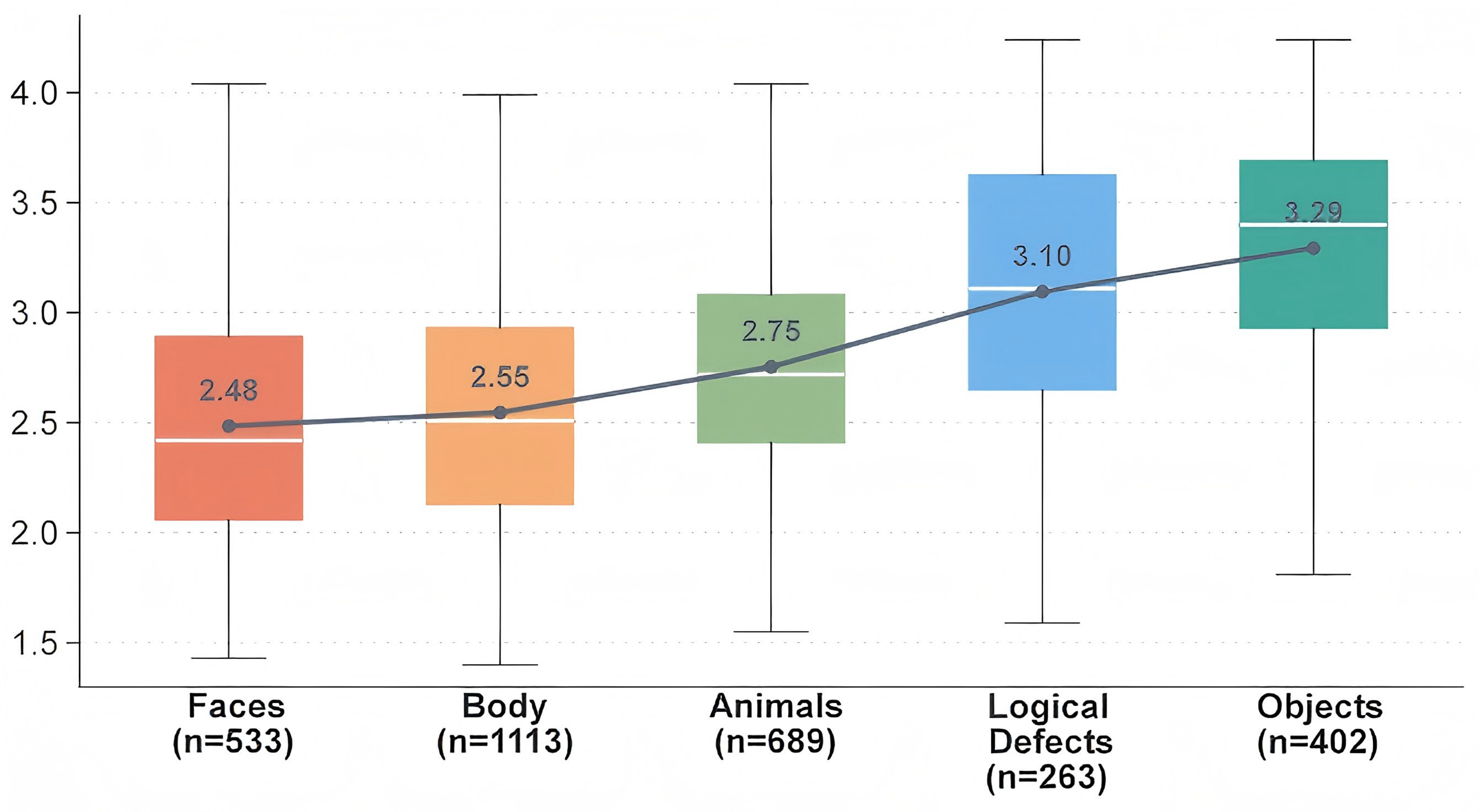}
\caption{\textbf{Quality score distribution across defect categories.} }
\Description{Box plots comparing image quality scores across five defect categories. The mean scores increase from face defects at 2.48 and body defects at 2.55 to animal defects at 2.75, logical defects at 3.10, and object defects at 3.29. Face and body defects therefore have the strongest negative effect on perceived image quality, while object and logical defects have relatively smaller effects.}
\label{fig:defect_category}
\vspace{-3ex}
\end{figure}

\begin{figure*}[t]
  \centering
  \includegraphics[width=0.85\textwidth ]{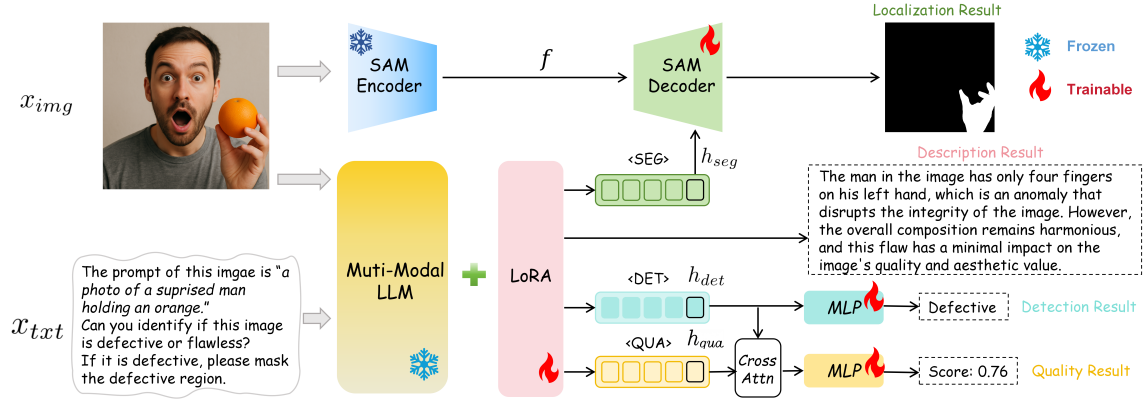}
\caption{\textbf{Overview of AGIDA architecture.} }
\Description{Architecture of AGIDA. An input image and text prompt are processed by a frozen multimodal large language model equipped with trainable LoRA modules. The SEG token guides a trainable SAM decoder, together with image features from a frozen SAM encoder, to produce a defect localization mask. The DET token is passed through a trainable detection head to predict whether the image is defective. The QUA token queries the defect representation through cross-attention and is mapped to a quality score. The multimodal language model also generates a textual explanation of the detected defect. Snowflake icons denote frozen components, and flame icons denote trainable components.}
\label{fig:method}
\end{figure*}

\subsection{Data Collection}

We employ a novel hybrid \textit{sample-and-generate} strategy, as illustrated in Figure~\ref{fig:construction}(a).

\textbf{Sampling from existing AGI datasets.} We curate images from existing AGI datasets~\cite{AGIQA-1K, AGIQA-3k, AGIQA-20K, RichHF-18K, aigciqa2023} along with their corresponding prompts. These datasets contain outputs from open-source models, with limited representation from proprietary systems.

\textbf{Augmentation via proprietary model generation.} To address this gap, we sample high-quality prompts from DiffusionDB~\cite{wang2022diffusiondb}, using GPT-4~\cite{GPT-4o} to verify coherence and filter ambiguous ones. From 20,000 vetted prompts, we generate images across 8 proprietary models: DALL-E 3~\cite{dalle3_openai}, Midjourney~\cite{midjourney}, FLUX~\cite{flux1_github}, Gemini~\cite{gemini2flash}, GPT-Image~\cite{gptimage1}, Ideogram~\cite{ideogram}, Kling~\cite{kling}, and Grok~\cite{xai2025grok3}. The resulting 4,000 prompts span seven categories, including Abstract Art (22.3\%), Landscapes (20.7\%), Animal (16.9\%), Indoor (13.8\%), City (12.4\%), Plant (8.1\%), and Human (6.0\%), ensuring broad semantic coverage. Model versions are detailed in the \textit{supplementary material}.

\subsection{Detection Annotation}
As shown in Figure~\ref{fig:construction}(b), the first stage focuses on binary defect detection. Annotators are presented with image-prompt pairs rather than images alone, enabling them to differentiate intentional stylistic choices from genuine defects. Each image is evaluated by at least four independent annotators. When unanimous agreement is reached, images are directly classified; for disputed cases, two additional annotators jointly review the images to establish final classifications. This two-tier approach achieves a 92.3\% initial agreement rate, ensuring that even minor defects are reliably identified, and yields 3,000 verified defective images and 1,000 flawless images.

\subsection{Localization, Explanation, and Score Annotation}

Following detection, we perform fine-grained annotation on defective images to localize defect regions, describe their characteristics, and assess overall image quality (Figure~\ref{fig:construction}(c)). This stage faces two challenges: defective images often contain multiple anomalies across different regions, increasing the risk of incomplete annotation; and annotators may over-interpret ambiguous regions as defective, leading to annotation noise.

\textbf{Step 1: Independent parallel annotation.} Each defective image is annotated by at least three independent annotators, who simultaneously provide: (i) segmentation masks delineating defective regions, (ii) textual descriptions covering defect type, spatial location, and impact on image quality, and (iii) an overall quality score reflecting the perceived image quality.

\textbf{Step 2: Annotation synthesis and validation.} Three-person expert panels review all annotations, discarding unreasonable ones through consensus. For validated annotations: (i) \textit{Mask integration}---reasonable masks are combined via union operations to capture the full extent of defective regions; (ii) \textit{Description synthesis}---GPT-4 coherently integrates multiple descriptions while preserving specific observations from each annotator; (iii) \textit{Score aggregation}---quality scores are averaged across annotators to obtain the final image-level score. Annotation interfaces, score criteria, and GPT-4 prompts are provided in the \textit{supplementary material}.

\subsection{Data Statistics }

Figure~\ref{fig:statistics} and Figure~\ref{fig:defect_category} present comprehensive statistics of AGIDefect-4K from multiple complementary perspectives.

\textbf{Defect characteristics.} The word cloud (Figure~\ref{fig:statistics}(a)) reveals a diverse spectrum of defect-related terms, spanning anatomical references (``finger,'' ``hand,'' ``leg''), quality descriptors (``unnatural,'' ``distorted''), and structural indicators (``missing,'' ``anomaly''), reflecting the broad coverage of our annotations.

\textbf{Localization complexity.} Figure~\ref{fig:statistics}(b) shows that over 85\% of defective regions occupy less than 5\% of the image area, highlighting that AGI defects are predominantly small and localized, making precise detection particularly challenging.

\textbf{Description richness.} Token lengths (Figure~\ref{fig:statistics}(c)) average 50--100 tokens and extend up to 200, indicating detailed annotations well beyond simple defect labels.

\textbf{Quality score analysis.} Figure~\ref{fig:statistics}(d) compares quality scores between defective and flawless images. Flawless images cluster around a higher mean with a compact distribution, while defective images exhibit lower scores with a wider spread. The clear separation validates that our quality annotations effectively capture the perceptual impact of defects, while the partial overlap reflects cases where minor defects have limited impact on overall quality.

\textbf{Defect category analysis.} To further investigate the relationship between defect types and perceived quality, we use GPT-4 to classify defects into five categories based on annotator descriptions (details in the \textit{supplementary material}). As shown in Figure~\ref{fig:defect_category}, body and face defects receive the lowest quality scores, indicating the most severe perceptual impact on overall image quality. This aligns with human perception: anatomical distortions in bodies and faces are more visually conspicuous than structural anomalies in objects or logical inconsistencies.

\section{AGIDA: A Baseline Model}

As illustrated in Figure~\ref{fig:method}, we present AGIDA (AGI Defect Assistant), a baseline framework that jointly performs defect detection, localization, explanation, and quality prediction.

\subsection{Architecture}

Drawing inspiration from LISA~\cite{lisa}, we extend the multimodal LLM paradigm with three task-specific tokens: \texttt{<DET>} for detection, \texttt{<SEG>} for segmentation, and \texttt{<QUA>} for quality prediction, enabling the model to simultaneously output detection results, localization masks, quality scores, and textual descriptions.

Given an input image $\mathbf{x}_{img}$ and a text prompt $\mathbf{x}_{txt}$, the multimodal LLM generates a textual response $\hat{\mathbf{y}}_{txt}$ containing defect descriptions:
\begin{equation}
    \hat{\mathbf{y}}_{txt} = \mathcal{F}_{\text{MLLM}}(\mathbf{x}_{img}, \mathbf{x}_{txt}).
\end{equation}

For defect detection, the \texttt{<DET>} token embedding $\mathbf{h}_{det}$ from the last hidden layer is passed through a detection head:
\begin{equation}
    \hat{\mathbf{D}} = \mathcal{F}_{det}(\mathbf{h}_{det}),
\end{equation}
where $\hat{\mathbf{D}} \in \{0, 1\}$ indicates flawless or defective status.

For defect localization, the \texttt{<SEG>} token embedding $\mathbf{h}_{seg}$ is projected through an MLP layer $\gamma$ and combined with visual features from a frozen SAM encoder:
\begin{equation}
\begin{aligned}
    \mathbf{f} &= \mathcal{F}_{enc}(\mathbf{x}_{img}), \\
    \hat{\mathbf{M}} &= \mathcal{F}_{dec}(\gamma(\mathbf{h}_{seg}), \mathbf{f}),
\end{aligned}
\end{equation}
where $\hat{\mathbf{M}}$ is the predicted segmentation mask highlighting defective regions.

For quality prediction, we enhance the \texttt{<QUA>} token embedding $\mathbf{h}_{qua}$ with defect-aware information from the \texttt{<DET>} token embedding $\mathbf{h}_{det}$ through cross-attention, and map the enhanced representation to a scalar quality score:
\begin{equation}
\begin{aligned}
    \mathbf{h}_{qua}' &=
    \operatorname{CrossAttn}
    \left(
        Q=\mathbf{h}_{qua},
        K=\mathbf{h}_{det},
        V=\mathbf{h}_{det}
    \right)
    + \mathbf{h}_{qua}, \\
    \hat{\mathbf{Q}} &= \mathcal{F}_{qua}(\mathbf{h}_{qua}').
\end{aligned}
\end{equation}
Here, $\mathbf{h}_{qua}$ serves as the query, while $\mathbf{h}_{det}$ serves as the key and value. This design injects defect-aware information into the quality representation, reflecting the intuition that the presence and severity of defects are important factors in perceived image quality.

\subsection{Training Objectives}

AGIDA is trained end-to-end with four complementary losses:
\begin{equation}
    \mathcal{L} = \lambda_{det} \mathcal{L}_{det} + \lambda_{mask} \mathcal{L}_{mask} + \lambda_{txt} \mathcal{L}_{txt} + \lambda_{qua} \mathcal{L}_{qua},
\end{equation}
where $\lambda_{det}$, $\lambda_{mask}$, $\lambda_{txt}$, and $\lambda_{qua}$ are weighting factors.

The detection loss employs cross-entropy for binary classification:
\begin{equation}
    \mathcal{L}_{det} = \text{CE}(\hat{\mathbf{D}}, \mathbf{D}).
\end{equation}

The mask loss combines binary cross-entropy and DICE loss:
\begin{equation}
    \mathcal{L}_{mask} = \lambda_{bce} \text{BCE}(\hat{\mathbf{M}}, \mathbf{M}) + \lambda_{dice} \text{DICE}(\hat{\mathbf{M}}, \mathbf{M}).
\end{equation}

The text generation loss uses auto-regressive cross-entropy:
\begin{equation}
    \mathcal{L}_{txt} = \text{CE}(\hat{\mathbf{y}}_{txt}, \mathbf{y}_{txt}).
\end{equation}

The quality loss employs mean squared error for score regression:
\begin{equation}
    \mathcal{L}_{qua} = \text{MSE}(\hat{\mathbf{Q}}, \mathbf{Q}).
\end{equation}

We employ LoRA~\cite{lora} for the multimodal LLM components while keeping the SAM encoder frozen. The detection head, cross-attention layer, quality regression head, projection layers, and SAM decoder are fully trainable.

\begin{table}[t!]
  \centering
  \caption{Comparison of detection and explanation performance on AGIDefect-4K. Best are in \textbf{bold}, second best are \underline{underlined}.}
  \label{tab:detection_explanation}
  \setlength{\tabcolsep}{2pt} 
  \resizebox{\linewidth}{!}{
  \begin{tabular}{@{}lcccccc@{}}
  \toprule
  \multirow{2}{*}{\textbf{Model}} & \multirow{2}{*}{\textbf{Params}} & \textbf{Detection} & \multicolumn{3}{c}{\textbf{Explanation}} & \multirow{2}{*}{\textbf{Rank}} \\
  \cmidrule(lr){3-3} \cmidrule(lr){4-6}
  & & \textbf{AUC} & Comp. & Prec. & \textbf{Mean} & \\
  \midrule
  \rowcolor{gray!10}
  \multicolumn{7}{c}{\textit{Closed-source Models}} \\
  \midrule
  Grok-4~\cite{grok_4} & NA & 0.29 & \underline{0.89} & 0.21 & 0.55 & 10/7 \\
  GPT-4o~\cite{GPT-4o} & NA & 0.31 & 0.54 & 0.38 & 0.46 & 6/8 \\
  Claude-Sonnet-4.6~\cite{claude_sonnet_4_6} & NA & 0.30 & 0.69 & 0.50 & 0.59 & 8/4 \\
  Qwen-VL-MAX~\cite{qwenvl} & NA & 0.38 & 0.51 & 0.35 & 0.43 & 3/9 \\
  Gemini-3.1-pro~\cite{gemini_3.1_pro} & NA & \underline{0.49} & \textbf{1.21} & \underline{0.93} & \textbf{1.07} & 2/1 \\
  \midrule
  \rowcolor{gray!10}
  \multicolumn{7}{c}{\textit{Open-source Models}} \\
  \midrule
  GLM-4.6V~\cite{glm} & 106B & 0.27 & 0.08 & 0.13 & 0.10 & 15/16 \\
  DeepSeek-VL2~\cite{deepseekvl2} & 27B-A4.5B & 0.28 & 0.24 & 0.13 & 0.19 & 14/15 \\
  LLaVA-v1.5~\cite{LLaVA} & 7B & 0.23 & 0.34 & 0.16 & 0.25 & 16/12 \\
  LLaVA-v1.5~\cite{LLaVA} & 13B & 0.29 & 0.31 & 0.16 & 0.24 & 10/13 \\
  Qwen3-VL~\cite{Qwen3-VL} & 235B-A22B & 0.29 & 0.60 & 0.21 & 0.41 & 10/10 \\
  mPLUG-Owl3~\cite{ye2024mplug} & 7B & 0.32 & 0.42 & 0.21 & 0.32 & 5/11 \\
  Llama-3.2-Vision~\cite{grattafiori2024llama} & 90B & 0.29 & 0.53 & 0.62 & 0.58 & 10/5 \\
  Kimi-K2.5~\cite{team2026kimi} & 1.1T & 0.31 & 0.72 & 0.40 & 0.56 & 6/6 \\
  Qwen3.5-Plus~\cite{qwen3.5} & 397B-A17B & 0.34 & 0.81 & 0.64 & 0.72 & 4/3 \\
  \midrule
  \rowcolor{gray!10}
  \multicolumn{7}{c}{\textit{Baseline \& Ours}} \\
  \midrule
  Qwen3-VL~\cite{Qwen3-VL} & 4B & 0.30 & 0.42 & 0.07 & 0.21 & 8/14 \\
  \rowcolor{blue!5}
  \textbf{AGIDA (Qwen3-VL)} & \textbf{4B} & \textbf{0.65} & \underline{0.89} & \textbf{0.95} & \underline{0.92} & 1/2 \\
  \bottomrule
  \end{tabular}
  }
  \vspace{-0.4cm}
  \end{table}
  
\begin{figure}[t]
\centering
\includegraphics[width=0.92\columnwidth]{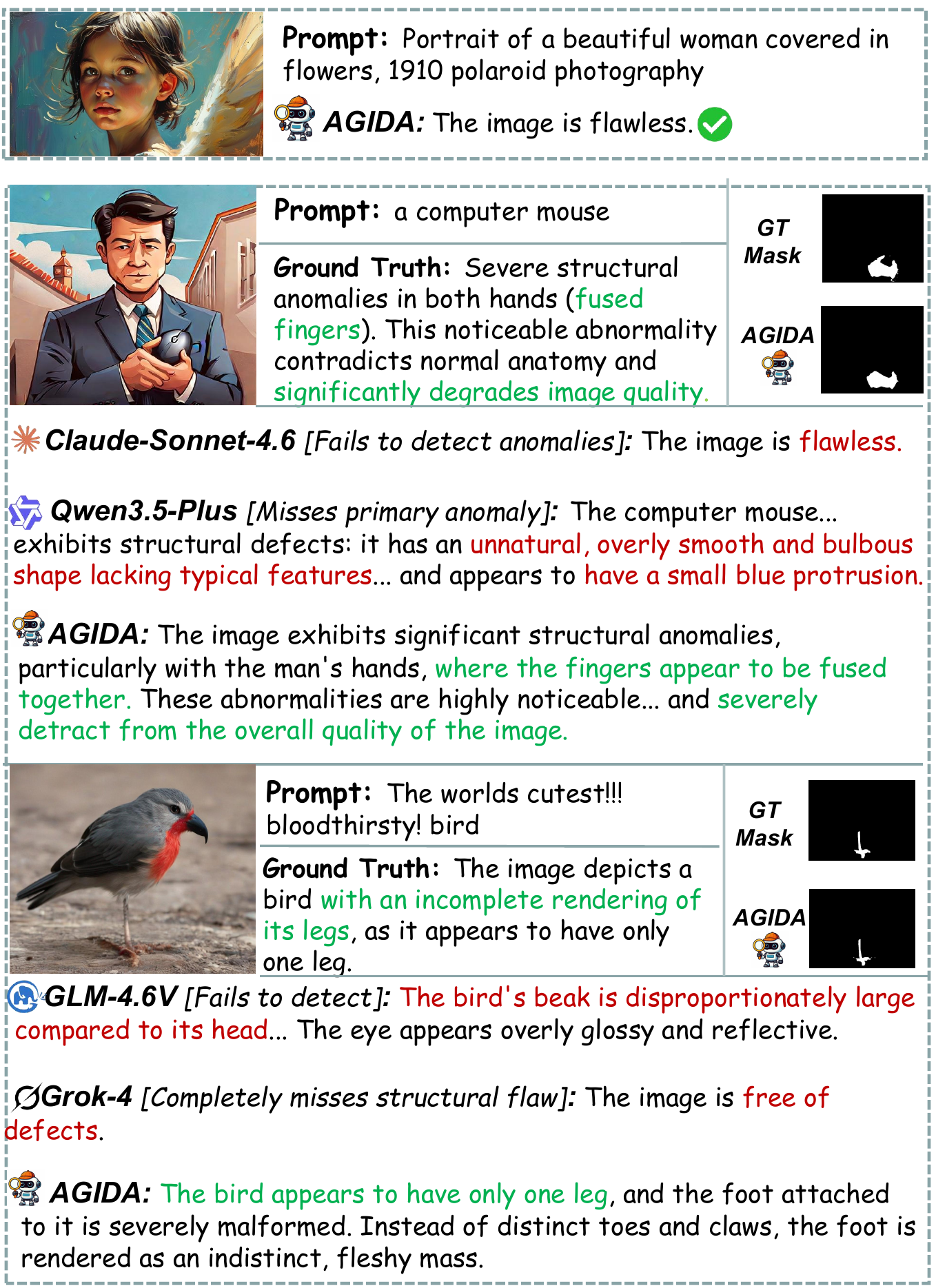}
\caption{Case studies. Best viewed when zoomed in.}
    \Description{Three qualitative examples comparing AGIDA with general-purpose multimodal models. In the first example, AGIDA correctly identifies a flawless portrait. In the second, the generated man has fused fingers: Claude-Sonnet-4.6 incorrectly declares the image flawless, and Qwen3.5-Plus focuses on the computer mouse, whereas AGIDA identifies the fused fingers and produces a mask aligned with the ground truth. In the third example, a bird has only one malformed leg: GLM-4.6V describes unrelated defects and Grok-4 misses the defect, while AGIDA correctly explains and localizes the malformed leg.}
\label{fig:case_study}
\end{figure}

\section{Experiments}

\subsection{Implementation Details}

We implement AGIDA based on Qwen3-VL-4B~\cite{Qwen3-VL}. All experiments are conducted on two NVIDIA RTX PRO 6000 GPUs (96GB). The per-device batch size is set to 1 with a gradient accumulation step of 4. We employ LoRA~\cite{lora} with rank $r=32$ and $\alpha=64$. The model is trained for 2 epochs using the AdamW optimizer with a weight decay of 0.1. The learning rate is initialized at $2\times10^{-5}$ with a cosine decay schedule. We partition AGIDefect-4K into non-overlapping training and testing sets using an 8:2 split. Results with additional backbones are provided in the \textit{supplementary material}.

\subsection{Evaluation Metrics}

\textbf{Detection.} We report AUC to measure overall detection capability.

\textbf{Localization.} We employ: (1) \textit{F1 score}; (2) \textit{GIoU} and \textit{CIoU}, where GIoU is the average of all per-image IoUs, and CIoU is defined by the cumulative intersection over the cumulative union.

\textbf{Explanation.} Following recent studies that leverage LLMs as reliable evaluators~\cite{wu2023q}, we adopt GPT-4o as an automated judge to assess generated defect descriptions against ground-truth annotations on two dimensions: (1) \textit{Completeness}---whether all defects are covered; (2) \textit{Preciseness}---whether the description is accurate and free of contradictions. Each dimension is rated from 0 (completely wrong) to 2 (fully correct), and we report the mean of both as the overall explanation score. Detailed evaluation prompts are provided in the \textit{supplementary material}.

\textbf{Quality prediction.} We evaluate quality score prediction using Spearman Rank Correlation Coefficient (SRCC) and Pearson Linear Correlation Coefficient (PLCC).

\begin{table}[t]
\centering
\begin{minipage}{0.5\columnwidth}
    \centering
    \caption{Defect localization results on AGIDefect-4K.}

    \label{tab:localization}
    \resizebox{\linewidth}{!}{%
    \begin{tabular}{@{}lccc@{}}
        \toprule
        \textbf{Model} & \textbf{F1$\uparrow$} & \textbf{GIOU$\uparrow$} & \textbf{CIOU$\uparrow$} \\
        \midrule
        PSCC-Net~\cite{pscc} & -- & -- & 0.095 \\
        EITLNet~\cite{EITLNet} & 0.317 & 0.178 & 0.193 \\
        IML-ViT~\cite{IML-ViT} & 0.110 & 0.182 & 0.152 \\
        LISA-13B~\cite{lisa} & -- & 0.226 & 0.167 \\
        \midrule
        \rowcolor{blue!5}
        \textbf{AGIDA (Qwen3-VL)} & \textbf{0.369} & \textbf{0.331} & \textbf{0.317} \\
        \bottomrule
    \end{tabular}%
    }
\end{minipage}
\hfill
  \begin{minipage}{0.48\columnwidth}
      \centering
      \caption{Quality assessment results on AGIDefect-4K.}
      \label{tab:quality}
      \resizebox{\linewidth}{!}{%
      \begin{tabular}{lcc}
          \toprule
          \textbf{Model} & \textbf{PLCC} & \textbf{SRCC} \\
          \midrule
          SF-IQA~\cite{SF-IQA} & 0.704 & 0.709 \\
          MOE-AGIQA~\cite{MOE-AGIQA} & 0.783 & 0.773 \\
          IPCE~\cite{IPCE} & 0.796 & 0.786 \\
          Diff-AGIQA~\cite{Diff-AGIQA} & 0.742 & 0.729 \\
          \midrule
          \rowcolor{blue!5}
          w/o CrossAttn & 0.791 & 0.787 \\
          \rowcolor{blue!5}
          \textbf{AGIDA (Qwen3-VL)} & \textbf{0.803} & \textbf{0.800} \\
          \bottomrule
      \end{tabular}%
      }
  \end{minipage}
\end{table}

\subsubsection{Detection and Explanation Performance}

Table~\ref{tab:detection_explanation} compares AGIDA with state-of-the-art MLLMs on AGIDefect-4K. All baseline MLLMs are evaluated in a zero-shot setting with Chain-of-Thought (CoT) prompts. Detailed prompt templates are provided in the \textit{supplementary material}.

\textbf{Detection.} Most zero-shot MLLMs achieve AUC scores between 0.23 and 0.38, indicating severe classification bias. Among closed-source models, Gemini-3.1-pro leads with 0.49 AUC, while most open-source models cluster around 0.29. AGIDA achieves the highest AUC of 0.65 with only 4B parameters, substantially outperforming all zero-shot baselines and demonstrating the effectiveness of task-specific fine-tuning.

\textbf{Explanation.} A consistent pattern emerges across models: completeness scores are generally higher than preciseness, suggesting that while models can partially identify defect presence, accurately characterizing defect types and details remains challenging. Gemini-3.1-pro ranks first in explanation with a mean of 1.07. Notably, AGIDA achieves the highest preciseness among all models and maintains competitive completeness (0.89), yielding the second-best overall explanation score.

\subsubsection{Localization Performance}
We adapt several relevant models for comparison by fine-tuning them on AGIDefect-4K, including state-of-the-art image manipulation detection and localization (IMDL) methods~\cite{pscc,EITLNet,IML-ViT}, and the reasoning segmentation model LISA~\cite{lisa}. As shown in Table~\ref{tab:localization}, IMDL methods show limited effectiveness on defect localization, highlighting the fundamental gap between detecting manipulations in real images and identifying generation defects in synthetic content. LISA achieves only modest performance, as it lacks binary detection capability and must be trained exclusively on defective images. AGIDA achieves substantial improvements across all metrics, demonstrating the effectiveness of joint detection-localization training.

\subsubsection{Quality Assessment Performance}
We compare AGIDA against state-of-the-art AGI quality assessment methods~\cite{SF-IQA,MOE-AGIQA,IPCE,Diff-AGIQA} fine-tuned on AGIDefect-4K. As shown in Table~\ref{tab:quality}, AGIDA achieves the highest PLCC and SRCC among all methods. Notably, incorporating cross-attention to inject defect detection features into the quality branch yields consistent improvements over the variant without it, validating that explicit defect awareness benefits quality prediction.

\subsection{Case Study}
Figure~\ref{fig:case_study} presents representative examples comparing AGIDA with state-of-the-art MLLMs. For the flawless image (top), AGIDA correctly identifies it as defect-free, avoiding false positives. For defective images, current MLLMs exhibit two common failure modes. The first is \textit{complete miss}: Claude-Sonnet-4.6 declares the hand-defect image ``flawless,'' and Grok-4 entirely overlooks the malformed bird leg. The second is \textit{misidentification}: Qwen3.5-Plus focuses on the mouse's shape rather than the actual fused-finger anomaly, while GLM-4.6V attributes issues to the bird's beak and eye instead of the missing leg. In contrast, AGIDA correctly identifies the fused fingers and the missing leg respectively, producing segmentation masks that closely align with ground truth. These results highlight the limitations of general-purpose MLLMs in fine-grained defect understanding. Additional visualization examples and failure case analyses are provided in the \textit{supplementary material}.

\section{Conclusion}
We presented AGIDefect-4K, a richly annotated dataset for comprehensive AGI defect diagnosis, featuring hierarchical annotations spanning detection, localization, explanation, and quality scoring across 4,000 images from 15 generative models. We further introduced AGIDA, a baseline framework that jointly addresses all four tasks, where a cross-attention mechanism bridges defect detection and quality prediction. Extensive benchmarking reveals that current MLLMs remain limited in fine-grained defect understanding, while dedicated defect modeling can benefit quality assessment. We hope AGIDefect-4K serves as a valuable resource for advancing AGI defect analysis and quality evaluation.

\begin{acks}
This work is supported by the National Natural Science Foundation of China under Grants 62471349, 625B2142, 62301378, 62501080, and 62171340; the Fundamental Research Funds for the Central Universities under Grants YJSJ25004 and QTZX25076; and partly by the China Postdoctoral Science Foundation under Grant 2024M762553. The authors thank the annotators from the BRAVE Lab at Xidian University for their valuable contributions to the dataset annotation.
\end{acks}

\bibliographystyle{ACM-Reference-Format}
\bibliography{sample-base}

\end{document}